\documentclass[conference]{IEEEtran}
\usepackage[dvipsnames]{xcolor}
\usepackage{array}
\usepackage{svg}
\usepackage{lineno}
\usepackage{listings}
\usepackage{longtable}
\usepackage{multirow}
\usepackage{booktabs} 
\usepackage{amsmath, amssymb, graphicx, hyperref}
\hypersetup{colorlinks=true, allcolors=blue}

\usepackage{orcidlink}

\title{Point, Detect, Count:\\ Multi-Task Medical Image Understanding with Instruction-Tuned Vision-Language Models
}

\author{
\IEEEauthorblockN{
Sushant Gautam\,\orcidlink{0000-0001-9232-2661}\IEEEauthorrefmark{1}\IEEEauthorrefmark{2},
Michael A. Riegler\,\orcidlink{0000-0002-3153-2064}\IEEEauthorrefmark{3},
Pål Halvorsen\,\orcidlink{0000-0003-2073-7029}\IEEEauthorrefmark{1}\IEEEauthorrefmark{2}
}
\IEEEauthorblockA{\IEEEauthorrefmark{1}Simula Metropolitan Center for Digital Engineering (SimulaMet), Norway}
\IEEEauthorblockA{\IEEEauthorrefmark{2}Oslo Metropolitan University (OsloMet), Norway}
\IEEEauthorblockA{\IEEEauthorrefmark{3}Simula Research Laboratory, Norway}
}

\usepackage{glossaries}
\makeglossaries
\newacronym{AI}{AI}{Artificial Intelligence}
\newacronym{bbox}{bbox}{Bounding Box}
\newacronym{GI}{GI}{Gastrointestinal}
\newacronym{IoU}{IoU}{Intersection over Union}
\newacronym{LLM}{LLM}{Large Language Model}
\newacronym{LoRA}{LoRA}{Low-Rank Adaptation}
\newacronym{MAE}{MAE}{Mean Absolute Error}
\newacronym{mAP}{mAP}{Mean Average Precision}
\newacronym{MSE}{MSE}{Mean Squared Error}
\newacronym{RMSE}{RMSE}{Root Mean Squared Error}
\newacronym{VLM}{VLM}{Vision-Language Model}

\begin{document}

\IEEEoverridecommandlockouts
\IEEEpubid{}  %

\maketitle
\begin{abstract}
We investigate fine-tuning \glspl{VLM} for multi-task medical image understanding, focusing on detection, localization, and counting of findings in medical images. Our objective is to evaluate whether instruction-tuned VLMs can simultaneously improve these tasks, with the goal of enhancing diagnostic accuracy and efficiency. Using MedMultiPoints, a multimodal dataset with annotations from endoscopy (polyps and instruments) and microscopy (sperm cells), we reformulate each task into instruction-based prompts suitable for vision-language reasoning. We fine-tune Qwen2.5-VL-7B-Instruct using \gls{LoRA} across multiple task combinations. Results show that multi-task training improves robustness and accuracy. For example, it reduces the Count \gls{MAE} and increases Matching Accuracy in the Counting + Pointing task. However, trade-offs emerge, such as more zero-case point predictions, indicating reduced reliability in edge cases despite overall performance gains. Our study highlights the potential of adapting general-purpose \glspl{VLM} to specialized medical tasks via prompt-driven fine-tuning. This approach mirrors clinical workflows, where radiologists simultaneously localize, count, and describe findings -- demonstrating how VLMs can learn composite diagnostic reasoning patterns. The model produces interpretable, structured outputs, offering a promising step toward explainable and versatile medical AI. Code, model weights, and scripts will be released for reproducibility at \href{https://github.com/simula/PointDetectCount}{https://github.com/simula/PointDetectCount}.
\end{abstract}


\section{Introduction}
\label{sec:intro}

Medical imaging comes with numerous challenges, such as detecting subtle abnormalities, processing images captured under varied conditions, and managing high data volumes, all of which complicate automated analysis~\cite{pmc_ncbi_nlm_nih_gov}. Among critical tasks are object detection (identifying lesions or instruments in images) and counting (quantifying occurrences, e.g., number of tumors or polyps), as these directly impact diagnoses and treatment planning. For example, in \gls{GI} endoscopy, detecting and counting polyps is vital since missed polyps (with reported miss rates of 14--30\%) can lead to cancer progression~\cite{Jha2019Dec}. Traditional computer vision methods have made progress in polyp detection accuracy~\cite{pmc_ncbi_nlm_nih_gov}, yet they often rely on single-modality inputs (images only) and focus on a single task. 
However, these approaches face two clinical limitations: 1) Although object detection models inherently provide counts by enumerating detected instances, their counting accuracy can be compromised in complex medical imaging scenarios, such as overlapping structures or small object sizes~\cite{Tyagi2023Jun}; 2) They lack natural language interfaces for query-based interaction that clinicians expect from decision support tools.
Moreover, these specialized models often lack the flexible, instruction-following capabilities of modern large language models, limiting their adaptability to nuanced clinical queries or combined analytical requests~\cite{Bai2024Oct}.

\begin{figure}
    \centering
    \includegraphics[trim={50mm 0mm 0mm 0mm}, clip, width=\linewidth]{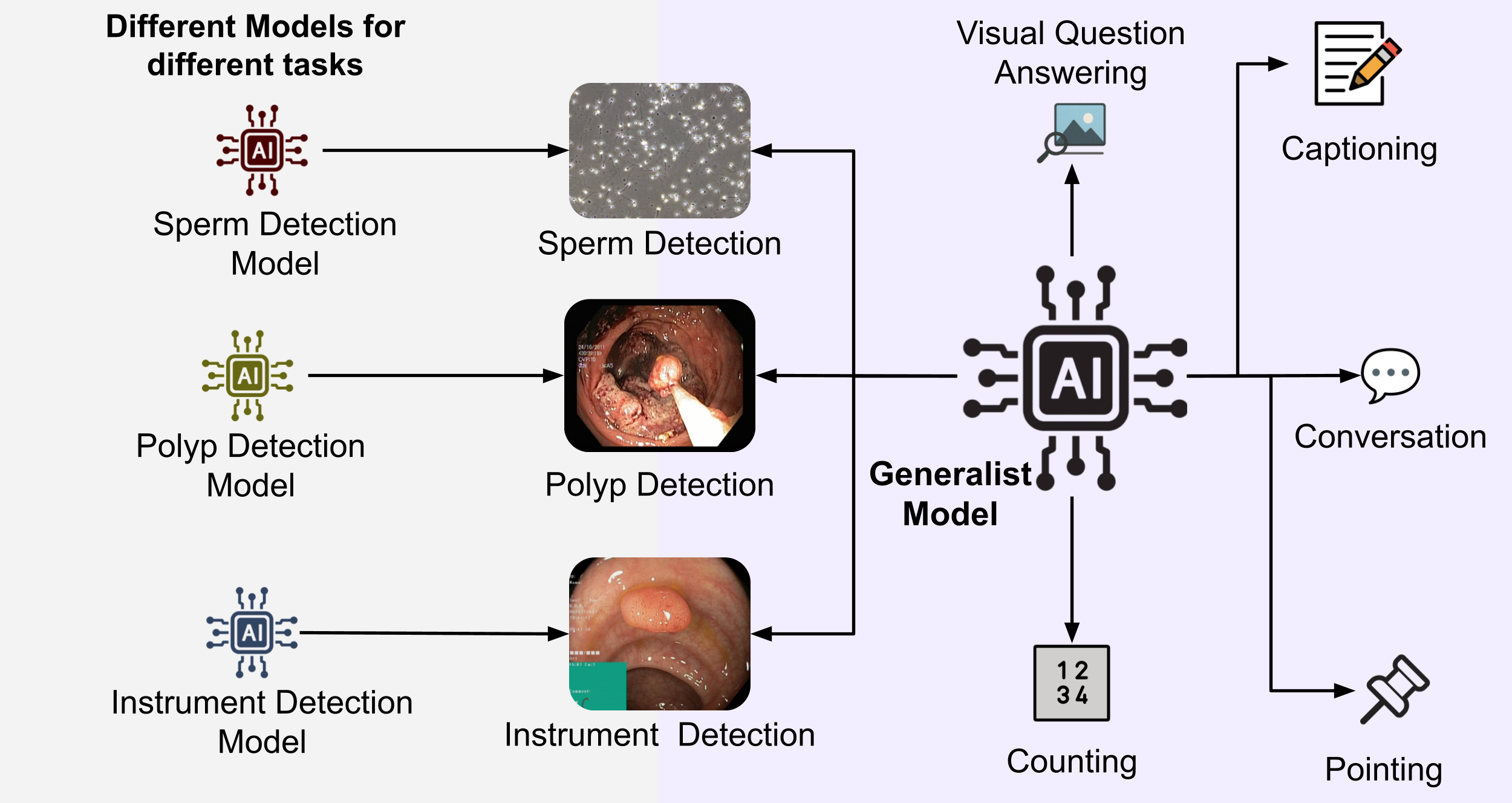}
    \caption{Illustration of different approaches to building multi-task AI systems}
    \label{fig:MultimodalMedicalAI}
\end{figure}

Multimodal \gls{AI} offers a promising approach to overcome these limitations by integrating information across multiple sources or tasks. By training models that handle images alongside auxiliary data or multiple objective outputs (e.g., bounding boxes, points and counts), we can leverage shared representations to improve performance~\cite{Zhang2021Mar}. 
Multimodal learning in this context, which involves using \glspl{VLM} with combined tasks, has the potential to enhance detection sensitivity and counting accuracy in medical imaging scenarios.
Figure ~\ref{fig:MultimodalMedicalAI} illustrates different approaches to building multimodal \gls{AI} systems, from using separate tools for each modality (left) to fully integrated generalist models (right).

Multimodal large models like Med-PaLM M can jointly interpret clinical text, images, and other data with one model~\cite{Tu2024Feb}. Recent advances in large multimodal models demonstrate the power of such integration. For instance, Google’s Med-PaLM Multimodal (Med-PaLM M) model can flexibly encode clinical language, medical images, and even genomic data with a single set of model weights~\cite{Tu2024Feb}. Notably, this generalist model achieved performance on a diverse benchmark of medical tasks that rivals or exceeds specialist models, highlighting the benefit of a multimodal, multi-task training paradigm~\cite{Tu2024Feb}.
Despite these advancements, there remains a gap in applying instruction-tuned VLMs to concurrently perform multiple tasks such as detection, localization, and counting within medical imaging. Our study addresses this gap by exploring the efficacy of such models in a multi-task learning framework.

Inspired by these developments, our work explores how a vision-language model can be fine-tuned for multiple objectives on a medical imaging dataset to improve object detection and counting. We hypothesize that training on combined tasks (bounding box localization, object counting, and pointing to object centers) will yield a model that outperforms single-task models on each individual task. We present a case study using the MedMultiPoints dataset of \gls{GI} endoscopy images and a large pre-trained multimodal model (Qwen2.5-VL-7B-Instruct), which we fine-tune for our specific task, to test this hypothesis.

 We demonstrate that a multimodal approach, one that fuses visual input with the model’s language reasoning and is fine-tuned on detection, counting, and pointing tasks, can significantly enhance performance in medical image analysis. 

The main contributions of our work are:
\begin{itemize}
    \item Formulating multiple vision-language tasks (detection, localization, and counting) from heterogeneous annotation sources and aligning them for instruction-tuned multimodal training.
    \item Demonstrating that multi-task fine-tuning improves both individual and joint-task performance, while revealing subtle failure modes through detailed error analysis.
    \item Providing comparative results between a public \gls{VLM} and our fine-tuned variant, with insights into task synergy, overfitting risks, and real-world reliability.
\end{itemize}

The following sections detail our approach and findings, emphasizing how multimodal \gls{AI} and multi-objective training lead to measurable improvements in detecting and counting clinical findings in images.

\section{Related Work}
\label{sec:related_work}

\subsection{Multimodal AI in Medical Imaging}

There is a growing recognition that many medical problems are inherently multimodal, involving images, textual reports, and other data~\cite{Tu2024Feb,AlSaad2024Sep,Kline2022Nov}. Multimodal learning approaches aim to combine these data sources to build richer models~\cite{Gao2020May}. In radiology, for example, the paradigm is shifting from image-only analysis to multimodal methods that integrate imaging with clinical text or other metadata~\cite{Yan2023Sep,Zhao2023Dec}. Large multimodal models like Med-PaLM M exemplify this trend by jointly learning from images and text and demonstrating strong generalization~\cite{Tu2024Feb}. Recent studies have demonstrated that vision-language pretraining enhances model robustness to imaging artifacts in endoscopy~\cite{Budathoki2025May}, which is crucial given the noisy nature of real-world medical video data.

Another line of research focuses on multimodal fusion of multiple imaging modalities (e.g., combining MRI and ultrasound) or integrating imaging with patient data to improve diagnostic accuracy~\cite{Tariq2025Jan,Azam2022May}. Studies consistently find that models leveraging multiple sources of information outperform those using a single modality~\cite{Huang2021Dec}. The benefits extend to multi-task scenarios as well: learning related tasks in parallel can act as a form of inductive transfer, where knowledge from one task helps improve others~\cite{Zhang2021Mar,Vandenhende2021Jan}. This multi-task learning approach enhances model robustness and generalization, especially in domains like medical imaging where data can be limited~\cite{Zhang2021Mar,Yu2024Jun}.

\subsection{Object Detection in Medical Images}
Object detection -- localizing and classifying findings in images -- is a well-studied problem in medical AI~\cite{Vidit2023Jan}. Traditional methods range from two-stage detectors (e.g., Faster R-CNN) to newer one-stage models like YOLO~\cite{Sultana2020Jun}. In \gls{GI} endoscopy, polyp detection has been a major focus~\cite{Durak2021Aug,Du2019Sep}. For instance, Wan et al. (2021) proposed an attention-enhanced YOLOv5 model for polyp detection~\cite{pmc_ncbi_nlm_nih_gov}. Their system automatically identifies polyps in colonoscopy frames, helping to reduce missed lesions and improve screening quality. Benchmark studies on polyp datasets such as Kvasir-SEG (1,000 polyp images with segmentation masks and bounding boxes) report detection precision and recall in the 0.88–0.91 range for advanced models~\cite{Jha2019Dec, pmc_ncbi_nlm_nih_gov}. These high-performing detectors underscore the feasibility of automated lesion detection.
However, most existing models are specialized networks trained solely for detection or segmentation, lacking the capability to address additional tasks such as counting the number of visible polyps. Recent research is bridging this gap. For example, Jha et al.~\cite{Jha2021Mar} utilized the Kvasir-SEG~\cite{Jha2019Dec} dataset for both detection (bounding box localization) and segmentation, highlighting that using the same dataset for multiple tasks encourages development of algorithms capable of both localization and segmentation.

\subsection{Counting and Pointing Tasks}
Counting objects in medical images is often as important as detecting them~\cite{Deitke2024Sep}. In pathology slides, one might count cells; in oncology, count metastases on scans; in endoscopy, count polyps. Specialized counting models (e.g., regression-based or density-estimation approaches) exist, but a trend is to incorporate counting into detection frameworks~\cite{Li2021Aug}. For example, a model might first detect all instances and then simply count them. Prior works on polyp detection have implicitly addressed counting by evaluating whether multiple polyps are all detected~\cite{pmc_ncbi_nlm_nih_gov}. Yet, explicit multi-objective models that output a count alongside detection results are less common in medical literature. 

The concept of pointing in \glspl{VLM} has emerged mainly in general \gls{AI}: given an image and a query, the model indicates the location (often as a point or bounding box) of the described object. In medical imaging, pointing could be used for referential localization (e.g., “show me the tumor”). Datasets with point annotations (e.g., centroids of objects) are valuable because point labels are easier to obtain than full segmentation masks while still conveying location. Point-based annotation has been suggested as a cost-effective method to reduce the intensive annotation effort typically associated with generating full segmentation masks or precise bounding boxes, while still providing valuable supervisory signals for localization tasks~\cite{Jha2021Mar}. The MedMultiPoints dataset provides such point labels for polyps, supporting a pointing task evaluation. 

Overall, the literature suggests that combining detection with auxiliary tasks (counting, pointing) in training could reinforce a model’s understanding of medical images, though concrete studies on multi-task training in medical object detection remain sparse. Our work builds on these insights, aiming to fill this gap by evaluating a multimodal model trained jointly on detection, counting, and pointing.

\subsection{Multimodal Vision-Language Models}
Large \glspl{VLM} have recently been applied to medical images. Models like Qwen-VL (7B parameters) and LLaVA are pre-trained on general images and aligned with language, enabling them to accept an image and produce a text output (descriptions, answers, etc.). Qwen-VL, developed by Alibaba, is a state-of-the-art multimodal LLM that uses a Transformer-based visual encoder (from OpenCLIP) coupled with a language model backend~\cite{Wang2024Sep}. Its variant, Qwen2.5-VL-7B-Instruct, is instruction-tuned for vision-language tasks~\cite{Bai2025Feb}.

These models can describe images or follow prompts about images, but out-of-the-box they are not specialized for medical data or structured outputs like bounding boxes. Nonetheless, their ability to understand visual content and respond flexibly makes them appealing for medical \gls{AI}, where a single model that can describe, detect, and answer questions about images would be highly useful~\cite{Tu2024Feb}. Early applications of \glspl{VLM} in medicine include radiology report generation and visual question answering (VQA) on medical images, with promising results when fine-tuned on domain-specific data. However, extending these capabilities to structured output tasks like precise object localization (points or bounding boxes) and counting directly from instruction prompts represents a more recent research thrust. 

This literature motivates our use of Qwen-VL as the base model: by fine-tuning it on a medical dataset with multi-task labels, we aim to imbue it with domain knowledge and the ability to output needed information (locations, counts) in text form. Building upon these foundational works, our study advances the field by focusing on the simultaneous application of instruction-tuned vision-language models for multi-task learning in medical imaging-specifically detection, localization, and counting. While recent efforts such as UMIT~\cite{UMIT}, LiteGPT~\cite{LiteGPT}, and MedViLaM~\cite{MedViLaM} have demonstrated progress in multi-task or multimodal medical AI, they typically target tasks like classification, report generation, or question answering in isolation or pairs. In contrast, our work uniquely reformulates all three structured tasks as instruction-driven prompts and fine-tunes a general-purpose VLM to handle them concurrently. This setup closely mimics clinical workflows, where radiologists often localize, count, and describe findings in tandem, thus pushing toward more holistic and versatile medical AI systems.

\section{The MedMultiPoints Dataset}

\begin{figure}[!t]
    \centering
    \includegraphics[width=\linewidth]{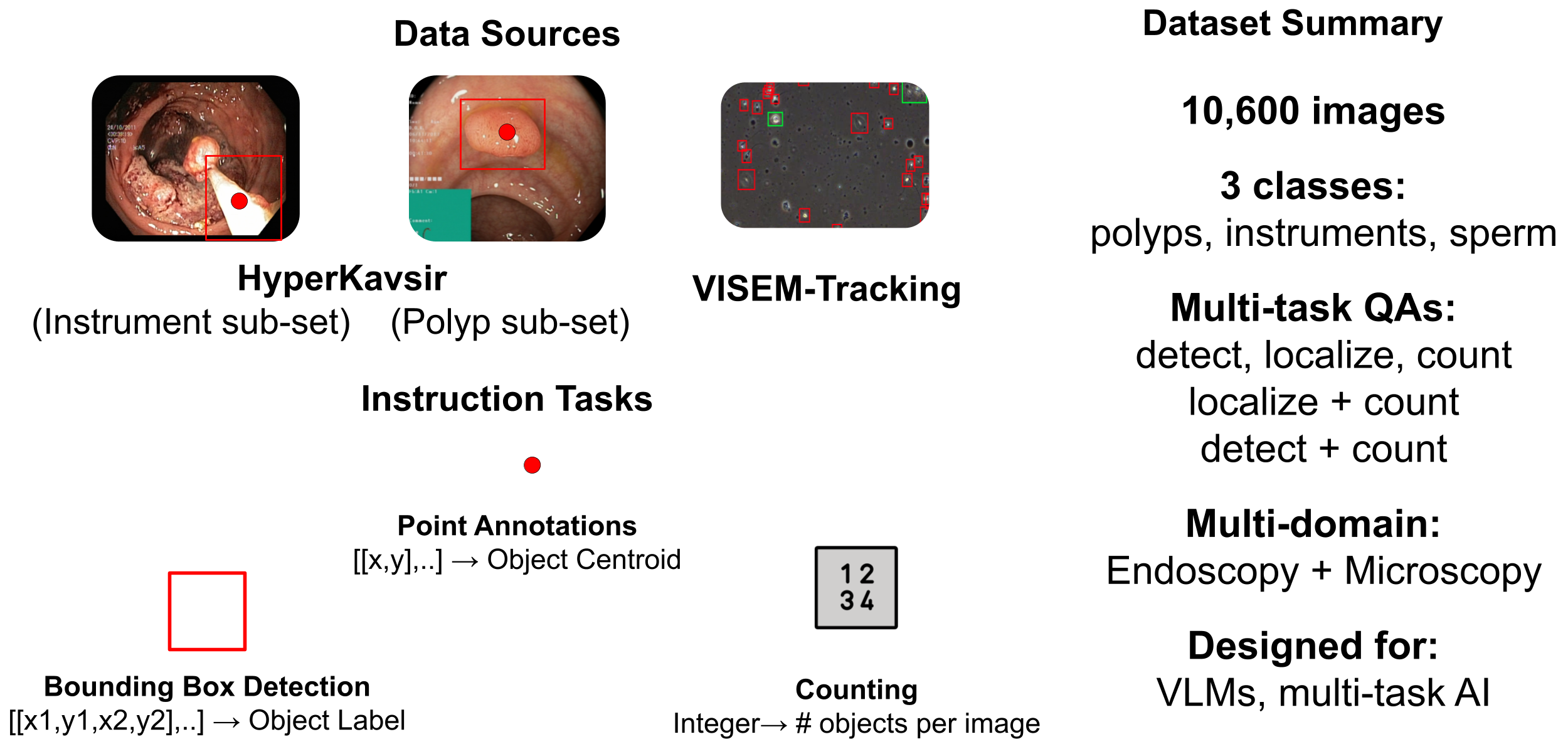}
    \caption{Overview of MedMultiPoints Dataset Structure and Annotation Types}
    \label{fig:DatasetOverview}
\end{figure}

The \textbf{MedMultiPoints} dataset, which we created for this paper and is openly available, is a multimodal medical imaging dataset designed for object detection, localization, and counting tasks.
Figure \ref{fig:DatasetOverview} provides a summary of the MedMultiPoints dataset, its sources, and annotation types.
It integrates annotations from three distinct sources: (1) {Polyp Bounding Box Annotations} from HyperKvasir~\cite{Jha2019Dec}, which focus on detecting and localizing polyps in Lower \gls{GI} endoscopic images; (2) {Instrument Bounding Box Annotations} from HyperKvasir, used for recognizing surgical instruments in endoscopic procedures; and (3) {Sperm Annotations} from microscopic imaging from VISEM-Tracking~\cite{Thambawita2023May}, which categorize sperm cells into normal, cluster, and pinhead classes. By incorporating data from both endoscopy and microscopy, MedMultiPoints provides a diverse dataset that supports various medical \gls{AI} applications.
The heterogeneity of MedMultiPoints, particularly its inclusion of diverse annotation types (bounding boxes, points, and counts) for the same or related image sets, is specifically designed to leverage the instruction-following and multi-task learning capabilities of VLMs~\cite{Zhu2025Jan}.

The dataset includes three primary annotation types: bounding boxes (\gls{bbox}), point annotations (\texttt{points}), and count annotations (\texttt{count}). Each type supports distinct tasks and contributes to the development of robust \gls{AI} models for medical imaging.

Bounding boxes are used to define the rectangular location of objects within an image. These are represented in JSON format as \texttt{[x1, y1, x2, y2]} and are especially useful for detecting objects such as polyps or surgical instruments. An example entry might be:
 \texttt{\{"bbox\_2d": [[x1, y1, x2, y2], ...], "label": "polpy"\}} 

Point annotations specify key locations within objects, such as the centroid of a polyp or the head of a sperm cell. These annotations are given as \texttt{[x, y]} coordinates. An example JSON entry could be:
 \texttt{\{"point\_2d": [[x, y], ...], "label": "polpy"\}} 

Count annotations provide the total number of instances of a specific object within an image, enabling training for tasks like density estimation or crowd counting. These are stored as integer values, for example:
 \texttt{\{"counts": 3, "label": "polpy"\}} 

This structured annotation approach enables \gls{AI} models to learn complementary tasks: object detection, localization, and counting -- with high accuracy. Each image is stored in a structured JSON format that integrates seamlessly into vision-language models and multimodal \gls{AI} pipelines.

The dataset, MedMultiPoints, comprises 10,600 images representing various clinical conditions. Images are resized proportionally to a maximum width of 700 pixels, if necessary. The dataset spans a range of scenarios, from images with no findings to those with multiple, densely packed objects. In addition to polyps, it includes surgical instruments captured during endoscopy and microscopic sperm images. This makes it one of the few datasets designed for multi-task detection in medical imaging.

Unlike conventional single-task datasets, MedMultiPoints is designed to support multi-objective training, where \gls{AI} models can simultaneously detect, count, and localize findings. This structure enables multimodal learning approaches, such as \gls{VLM}s and multi-task neural networks, which leverage overlapping representations across different tasks. The dataset can serve as a benchmark for object detection and multimodal \gls{AI} applications in medical imaging.

\section{Methodology}
\label{sec:methodology}
\subsection{Model Architecture}
\begin{figure}[ht]
    \centering
    \includegraphics[width=\linewidth]{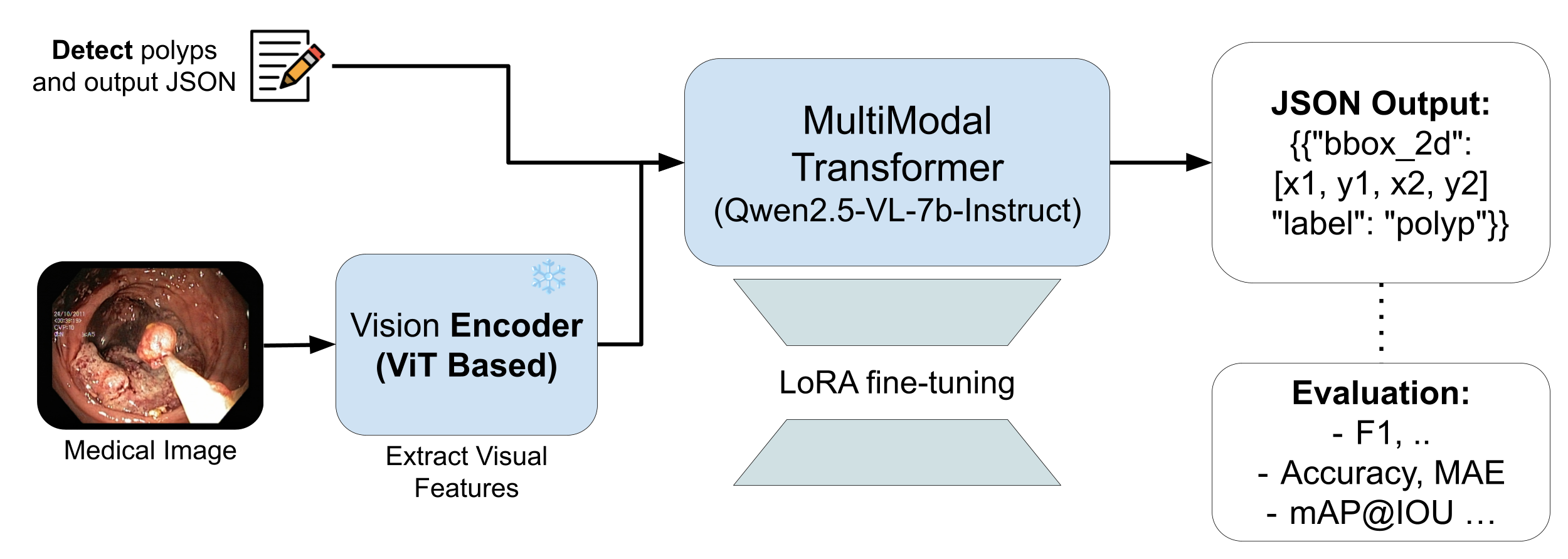}
    \caption{Overview of Model Architecture}
    \label{fig:Architecture}
\end{figure}

Our model is based on Qwen2.5-VL-7B-Instruct~\cite{Bai2025Feb}, a large multimodal Transformer model that can process visual and textual inputs and generate text outputs. The architecture comprises a vision encoder (a ViT-based model) that produces image features, and a language model decoder that generates responses.
Its instruction-tuned nature makes it particularly amenable to our approach, where tasks are defined via natural language prompts requiring specific, structured textual outputs~\cite{Wei2021Sep}.
By default, Qwen2.5-VL is trained for tasks like image captioning and question answering and has been enhanced to support object grounding. We adapt this model to our tasks by formulating detection, counting, and pointing as a combined instructed output problem.
The overview of the architecture is shown in Figure~\ref{fig:Architecture}.

Specifically, we prompt the model with the image (as input to the visual module) and a task instruction to produce the desired output. For example, a training prompt might be: \textit{"Return bounding boxes for each sperm in the image and the total count"}. The expected output for this particular example should be a valid JSON string encoding the bounding boxes and count information.
For pointing, we experimented with prompting the model to output point coordinates independently rather than having it to output bounding boxes and then extract centroids. 

\subsection{LoRA Fine-Tuning}
To efficiently adapt Qwen2.5-VL-7B-Instruct without updating all seven billion parameters, we employed \gls{LoRA}~\cite{hu2022lora}. \gls{LoRA} introduces small trainable rank-decomposition matrices into the model’s linear layers, enabling parameter-efficient fine-tuning while keeping the original pre-trained weights frozen. This reduces the number of trainable parameters from billions to just a few million, making training feasible on a single GPU.
This parameter efficiency is particularly crucial in medical AI, where small annotated datasets (n=10,600 here vs. millions in natural images) make full fine-tuning prone to catastrophic forgetting of the model's original visual-language alignment.
This strategy allowed stable and effective fine-tuning even with a modest dataset size, aligning with best practices in multimodal model adaptation.
In our setup, only the \gls{LLM} component was fine-tuned, while the image encoder (Vision Transformer) remained frozen to reduce computational costs and overfitting risks, and to leverage the strong pretrained visual features already captured by the encoder. \gls{LoRA} adapters were applied to all linear layers in the LLM, excluding the final language modeling head. This exclusion is common, as adapting earlier layers suffices for stylistic and task alignment while preserving the pretrained head's generative capacity~\cite{hu2022lora}.
We used a \gls{LoRA} rank of 16 to provide greater adaptation capacity for the model’s multi-task medical imaging objectives, improving performance without significant overhead. We also carefully tuned the scaling factors to maintain stability and preserve the model’s pre-trained capabilities. This choice balances the trade-off between model complexity and computational efficiency, ensuring that the fine-tuned model remains practical for real-world clinical applications.

\subsection{Training Procedure}
For each image, we constructed five distinct instruction-response pairs, one for each of the five tasks. Each instruction comprised a templated task description combined with a question randomly selected from a predefined set of question templates corresponding to that task. This template-based design introduced significant linguistic diversity, ensuring that the model was exposed to a wide variety of input phrasings. The response was a stringified JSON dictionary containing the model’s predictions, including a list of bounding box coordinates, point coordinates, and object counts as appropriate for the task. All tasks were combined under a unified training objective, prompting the model to simultaneously localize and count objects across multimodal queries.
\begin{figure}[ht]
    \centering
\includegraphics[width=\linewidth]{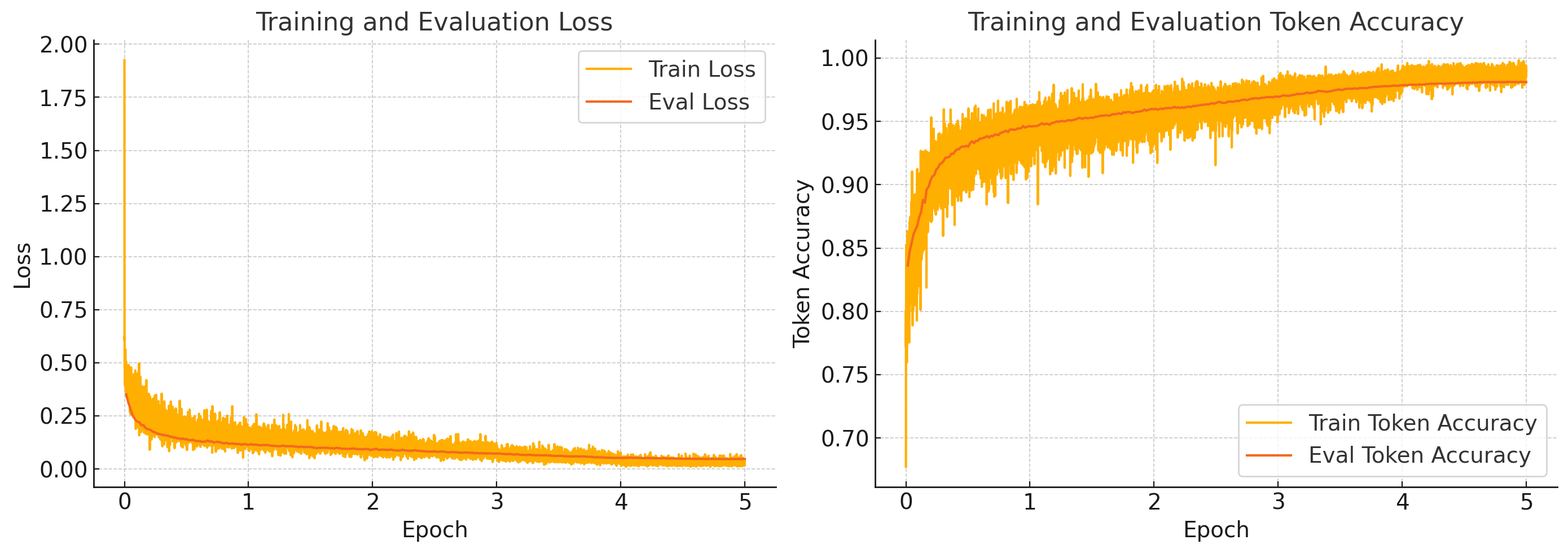}
    \caption{Training and evaluation curves over five epochs. Left: Loss convergence. Right: Token accuracy stabilization, indicating effective generalization.}
    \label{fig:training_loss}
\end{figure}
Training was done for five epochs on the training set using the AdamW optimizer on a single NVIDIA A100 80GB.
The number of epochs was based on empirical observations of loss convergence and token accuracy stabilization as shown in Figure \ref{fig:training_loss}, following best practices for instruction tuning of pretrained vision-language models.
The learning rate was set to 2e-4 for the \gls{LoRA} parameters, with all other weights frozen. We leveraged gradient accumulation to handle a batch size of four per step, evaluating every 200 steps.

Since the model was trained to generate outputs in a structured JSON format containing coordinates and counts, we employed a parser to extract these elements from the predicted text. The training objective was the standard language modeling loss (cross-entropy over tokens), which ensured that the model learned to produce syntactically valid and semantically correct JSON outputs.
This unified text-based output, despite encoding diverse information types like coordinates and counts, allows the VLM to be trained with a standard language modeling objective. To evaluate performance, we parsed the model’s output and aligned it with the ground truth annotations. For instance, if the model returned two bounding boxes while the ground truth contained three polyps, it incurred a penalty for the missed detection. This enabled task-specific evaluation while maintaining a unified, text-based output space.

\subsection{Evaluation Metrics}

We evaluated our model across five multimodal annotation prediction tasks using task-specific metrics. For counting tasks, we report \gls{MAE} and \gls{MSE} to capture both average and large deviations in predicted object counts. For pointing tasks, we use Point \gls{MAE} and \gls{RMSE} to measure spatial localization error, Matching Accuracy to indicate the fraction of predicted points within 10 pixels of ground truth, and Zero-case Points to count failure cases with no predictions. For bounding box detection tasks, we report \gls{mAP}, \gls{mAP}@50, \gls{mAP}@75 to assess detection quality, along with average \gls{IoU} as a measure of box alignment with ground truth. Together, these metrics provide a comprehensive view of accuracy, robustness, and spatial precision across annotation types.

Additionally, we qualitatively assess the textual outputs for correctness and clarity, since the model’s results are in human-readable form. The fine-tuned model is compared against the public Qwen2.5-VL-7B-Instruct model without fine-tuning, which we use as a baseline to generate responses on the same evaluation set. The baseline is prompted in an identical way (asking it to identify and count polyps) even though it was not specifically trained for this; this tests its zero-shot abilities. All evaluations were performed on the held-out test set of 500 samples.

\section{Results}
\label{sec:experiments}
\begin{table}[!bt]
\centering
\caption{
\textbf{Performance comparison of multimodal annotation prediction tasks.} Lower values are better for MAE, MSE, RMSE, and Zero-case Points; higher values are better for mAP, IoU, and Matching Accuracy.
}
\begin{tabular}{|p{2.2cm}|l|r|r|}
\hline
\textbf{Task} & \textbf{Metric} & \textbf{Qwen-public} & \textbf{Ours} \\
\hline
\multirow{2}{2.2cm}{\centering \textbf{Counting Only} \\ (n=105)} 
& Count MAE & 9.86 & 0.26 \\
& Count MSE & 389.04 & 2.62 \\
\hline
\multirow{4}{2.2cm}{\centering \textbf{Pointing Only} \\ (n=91)} 
& Point MAE & 52.50 & 1.24 \\
& Point RMSE & 57.04 & 1.64 \\
& Matching Accuracy & 0.43 & 0.99 \\
& Zero-case Points & 68 & 0 \\
\hline
\multirow{4}{2.2cm}{\centering \textbf{Bounding Box}\\\textbf{Detection} \\ (n=107)} 
& mAP & 0.01 & 0.85 \\
& mAP@50 & 0.01 & 0.95 \\
& mAP@75 & 0.01 & 0.88 \\
& IoU (avg) & 0.21 & 0.97 \\
\hline
\multirow{6}{2.2cm}{\centering \textbf{Counting +} \\ \textbf{Pointing} \\ (n=98)} 
& Count MAE & 6.70 & 1.52 \\
& Count MSE & 156.09 & 18.19 \\
& Point MAE & 92.42 & 17.78 \\
& Point RMSE & 97.44 & 23.40 \\
& Matching Accuracy & 0.25 & 0.91 \\
& Zero-case Points & 37 & 29 \\
\hline
\multirow{6}{2.2cm}{\centering \textbf{Counting +} \\ \textbf{Bounding} \\ (n=99)} 
& Count MAE & 9.63 & 1.37 \\
& Count MSE & 366.54 & 43.03 \\
& mAP & 0.00 & 0.80 \\
& mAP@50 & 0.01 & 0.92 \\
& mAP@75 & 0.00 & 0.82 \\
& IoU (avg) & 0.21 & 0.89 \\
\hline
\end{tabular}
\label{tab:model_comparison}
\end{table}

We evaluate three fundamental clinical competencies: 1) Detection (Can it find all relevant objects?), 2) Quantification (Can it count them accurately?), and 3) Spatial Reasoning (Can it precisely localize findings?).
Table~\ref{tab:model_comparison} compares a fine-tuned Qwen2.5-VL-7B-Instruct model ("Ours") with the public Qwen2.5-VL-7B-Instruct checkpoint across five annotation tasks, including bounding box detection, point localization, object counting, and their combinations. Key metrics include \gls{mAP} and \gls{IoU} for box quality, \gls{MAE}/\gls{MSE} for counting and localization errors, and Matching Accuracy, which measures how many predicted points fall within ten pixels of the ground truth. Zero-case Points count samples where no predictions were made, revealing model failures.

\subsection{Detection Performance}
The fine-tuned model significantly outperformed the public Qwen2.5-VL-7B-Instruct checkpoint in object detection tasks. As shown in Table~\ref{tab:model_comparison}, our model achieved an overall \gls{mAP} of {0.85}, compared to only {0.01} for the public model. At the \gls{IoU} threshold of 0.5, our model achieved {mAP@50 = 0.95}, and at a stricter threshold of 0.75, it maintained strong performance with {mAP@75 = 0.88}. Furthermore, the average \gls{IoU} increased dramatically from {0.21} to {0.97}, indicating precise spatial alignment. The public model frequently failed to return any valid bounding boxes, whereas the fine-tuned model consistently produced accurate detections across all samples.

\subsection{Counting Performance}
Counting accuracy was also greatly improved after fine-tuning. On the standalone \textit{Counting Only} task, our model achieved a remarkably low \gls{MAE} of {0.26} and \gls{MSE} of {2.62}, compared to {MAE = 9.86} and {MSE = 389.04} for the public Qwen2.5-VL-7B-Instruct checkpoint. Even under multi-task conditions, the fine-tuned model retained strong performance. For instance, in the \textit{Counting + Pointing} task, the model achieved {MAE = 1.52} and {MSE = 18.19}; in the \textit{Counting + Bounding} task, it maintained {MAE = 1.37} and {MSE = 43.03}. These results suggest robust enumeration capabilities even when other spatial annotations are predicted simultaneously.

\subsection{Pointing (Localization) Performance}
The localization task also showed substantial improvements. In the \textit{Pointing Only} setting, our model achieved a point \gls{MAE} of {1.24} and \gls{RMSE} of {1.64}, a significant reduction from the public model's errors ({MAE = 52.50}, {RMSE = 57.04}). Matching Accuracy, which counts predictions within 10 pixels of ground truth, improved from {0.43} to {0.99}, while the number of zero-case samples dropped from {68} to {0}. Under multi-task conditions like \textit{Counting + Pointing}, the fine-tuned model maintained high accuracy with MAE = 17.78, RMSE = 23.40, and Matching Accuracy = 0.91, again outperforming the public checkpoint on all fronts.

\subsection{Performance Gains vs. Prediction Stability}
Despite the significant performance improvements observed by epoch~5, some concerns emerge around overfitting and prediction reliability in complex multimodal tasks.
For instance, in the \textit{Counting + Pointing} task, while the Count \gls{MAE} drops from 8.37 (epoch 1, not shown) to 1.52 (epoch 5), the number of zero-case point predictions rises sharply from 3 to 29.
This suggests that the model, despite improving on average error, is failing to produce outputs on harder examples.
A likely cause is overfitting: the model appears to prioritize easier instances to minimize loss, while becoming overly cautious or under-confident on difficult ones.
This behavior indicates poor generalization, masked by improvements in global metrics.
Similarly, in the \textit{Counting Only} task, although the Count \gls{MAE} improves from 0.39 to 0.26, the Count \gls{MSE} increases from 1.25 to 2.62, reflecting higher error variance likely driven by outliers.
Together, these patterns highlight the limitations of relying solely on aggregate metrics like \gls{MAE} and emphasize the need for complementary evaluations that can reveal instability and failure modes in multimodal settings.
These results highlight a potential trade-off: while extended training (up to epoch five) enhances accuracy and matching quality, it may reduce the model’s robustness in edge cases or introduce overconfidence, warranting careful monitoring during continued fine-tuning.

\subsection{Comparative Analysis}
The public Qwen2.5-VL-7B-Instruct model, without fine-tuning, showed consistently poor performance across tasks and failed to handle complex multimodal annotations, highlighting the necessity of task-specific adaptation.
We chose this baseline due to its architectural parity with our fine-tuned model and its accessibility for reproducibility. Furthermore, existing medical VLMs such as Med-PaLM M or LLaVA-Med are not designed for object detection or boundary prediction tasks, limiting their comparability in our multi-task setup.

Across all tasks, the fine-tuned Qwen2.5-VL-7B-Instruct model (“Ours") consistently and significantly outperformed the publicly available Qwen2.5-VL-7B-Instruct checkpoint. 
The most striking improvements were seen in detection and pointing tasks, where metrics improved by over an order of magnitude. The public model often failed to produce outputs, especially in spatial prediction tasks, as indicated by high zero-case counts and near-zero mAP scores. In contrast, our model delivered structured, accurate predictions across counting, pointing, and bounding box detection, both individually and in combination.

While slight degradation was observed in multi-task scenarios compared to their single-task counterparts, especially in counting, our model still outperformed the public model by large margins. This trade-off may result from token-level competition and the increased complexity of multimodal formatting. Nonetheless, the results confirm that fine-tuning enables a general-purpose vision-language model like Qwen2.5-VL-7B-Instruct to adapt effectively for multimodal medical annotation tasks, delivering high precision in structured outputs suitable for downstream \gls{AI} pipelines.
These results suggest that instruction-tuned VLMs can effectively handle complex, multi-faceted tasks in medical imaging, potentially reducing the need for multiple specialized models and simplifying the deployment of AI in clinical workflows.

\section{Discussion}
\label{sec:discussion}

The experimental findings demonstrate that our fine-tuned Qwen2.5-VL-7B-Instruct model achieves significant performance gains over the public checkpoint across multiple annotation tasks. However, the results also expose nuanced trade-offs, particularly related to overfitting and reliability in edge cases. Below, we discuss the implications and limitations uncovered in our evaluation.

\subsection{Task-Specific Improvements and Trade-Offs}

Across all tasks-bounding box detection, point localization, object counting, and their combinations -- our model consistently outperforms the baseline. Notably, in the \textit{Counting + Pointing} task, Count \gls{MAE} improves markedly from 8.37 (epoch~1) to 1.52 (epoch~5), indicating the model’s growing ability to infer object quantities. However, the concurrent rise in zero-case point predictions from 3 to 29 suggests a troubling regression in robustness: while the model becomes more accurate on average, it increasingly fails to generate outputs in more challenging samples. This points to a trade-off between fine-tuned precision and general coverage.
In clinical contexts, such omissions could result in missed findings -- especially problematic for early diagnosis or real-time decision support. These zero-case outputs highlight the model’s sensitivity to hard samples and suggest a need for mechanisms like uncertainty estimation, abstention-based prediction, or fallback triggers to handle uncertain or ambiguous cases more gracefully.

A similar trend is observed in the \textit{Counting Only} task, where the Count \gls{MAE} decreases from 0.39 to 0.26, yet Count \gls{MSE} increases from 1.25 to 2.62, revealing that a few high-error predictions dominate the variance. These findings indicate that although extended training improves overall metrics, it may inadvertently promote overconfidence or exacerbate errors on edge cases.

\subsection{Effects of Multi-Task Fine-Tuning}

Multi-task training appears to yield synergistic effects. By jointly learning to detect, count, and localize, the model develops stronger internal visual representations. This is especially evident in the \textit{Counting + Pointing} and \textit{Box + Count} tasks, where improvements in one metric often coincide with better performance in the other. For example, a lower Count \gls{MAE} typically aligns with higher Matching Accuracy, implying that accurate counting helps reinforce correct spatial reasoning and vice versa.

However, the fine-tuning trajectory also reveals potential pitfalls: aggressive optimization toward one task (e.g., minimizing counting error) can cause degradation in complementary metrics (e.g., missing point predictions). This suggests a delicate balance in loss weighting is essential to avoid skewed task prioritization.

\subsection{Prediction Reliability and Failure Modes}

The increased occurrence of zero-case point predictions and elevated error variance highlight the need for better calibration. Our model occasionally abstains from making predictions in difficult scenarios, possibly due to uncertainty or over-regularization. This behavior underscores the importance of incorporating uncertainty modeling or fallback mechanisms, such as confidence thresholds or auxiliary heads, to flag or recover from failure cases.

Additionally, although textual outputs offer interpretability, they introduce formatting fragility. Minor variations in punctuation or phrasing can hinder automated evaluation, requiring careful prompt engineering and parsing. Constrained decoding or enforced output schemas could enhance consistency without sacrificing flexibility.

\subsection{Implications for Clinical Deployment}

The performance trajectory across epochs suggests that monitoring beyond aggregate accuracy is essential. Metrics like zero-case predictions, Matching Accuracy, and error variance provide critical insight into model robustness, especially in clinical contexts where missed detections carry high risk.

Despite these challenges, the fine-tuned Qwen2.5-VL-7B-Instruct model offers promising capabilities: the ability to perform multiple vision-language tasks with a single unified model, improved structured outputs, and competitive performance with minimal data. Compared to traditional models that rely on rigid outputs (e.g., YOLO), our approach allows for richer image understanding and explanation in natural language -- albeit with trade-offs in parsing and real-time suitability.
The structured outputs, delivered in a machine-readable JSON format, facilitate downstream integration into clinical systems, such as report generation or quality assurance pipelines. However, we acknowledge that our dataset primarily focuses on GI imaging and spermogram, and generalizing the model to more modalities like CT or MRI would require targeted adaptation, which we leave as future work.

Future work could focus on dynamic loss balancing, error-aware learning strategies, and domain transferability to other imaging modalities. Overall, while our results affirm the power of instruction-tuned multimodal models, they also emphasize the need for thoughtful deployment practices and failure-aware evaluation protocols.

\section{Conclusion}
This study investigates the effectiveness of fine-tuning a large vision-language model on diverse, structured tasks in medical image understanding. Using \gls{LoRA}-based adaptation, we trained the model on a composite dataset built from multiple annotation sources, covering object detection (\gls{bbox}), localization (pointing), counting, and their combinations. The goal was to explore how multi-objective learning affects the performance of a multimodal model when applied to real-world, clinically relevant tasks.

Our experiments reveal that strategic fine-tuning, especially under a multi-task fusion setup, unlocks substantial performance gains over the public checkpoint, highlighting the power of jointly optimizing related tasks. For instance, in the \textit{Counting + Pointing} task, fine-tuning reduced count error while improving point-level matching -- showing that the joint learning of related objectives enables the model to generalize better. However, we also observed trade-offs, such as increased failure rates in harder examples (e.g., zero-case predictions), emphasizing the need for balanced optimization and robustness checks. This careful balance is crucial for translating promising research findings into reliable clinical tools.
Beyond performance, the fine-tuned model retains the interpretability strengths of \gls{VLM}s, producing structured, readable outputs that could assist clinical workflows. This is particularly relevant for diagnostic contexts where both quantitative precision (e.g., object counts) and qualitative descriptions are needed.

Looking forward, our approach sets the stage for generalizable, multi-task medical \gls{AI} systems. Future directions include extending this framework to other pathologies and imaging modalities, integrating structured patient metadata for richer prompts, and enhancing output consistency through constrained decoding or syntax-aware prompting. Additionally, exploring real-time constraints and uncertainty estimation will be crucial for safe deployment in clinical environments.

In summary, this work affirms the potential of vision-language foundation models when carefully adapted through task fusion and domain-specific fine-tuning. By demonstrating the feasibility of multi-task learning in medical imaging, our study contributes to the ongoing efforts to develop more versatile and efficient AI tools that can seamlessly integrate into various aspects of patient care.

\section*{Acknowledgement}
This work has benefited from the Experimental Infrastructure for Exploration of Exascale Computing (eX3), which is financially supported by the Research Council of Norway under contract 270053.

\section*{Use of AI Disclosure}
Various AI/LLM tools were used to draft structure, improve language and clarity. All content has been carefully manually reviewed, verified, and finalized by the authors.

\bibliographystyle{ieeetr}
\bibliography{main}

\end{document}